# Evolving Classifiers: Methods for Incremental Learning

Gregory Hulley and Tshilidzi Marwala

School of Electrical and Information Engineering

University of the Witwatersrand

Private Bag x3

Wits, 2050

Republic of South Africa

E-mail: greghul@icon.co.za

tshilidzi.marwala@wits.ac.za

**Abstract.**

The ability of a classifier to take on new information and classes by evolving the classifier without it having to be fully retrained is known as incremental learning. Incremental learning has been successfully applied to many classification problems, where the data is changing and is not all available at once. In this paper there is a comparison between Learn++, which is one of the most recent incremental learning algorithms, and the new proposed method of Incremental Learning Using Genetic Algorithm (ILUGA). Learn++ has shown good incremental learning capabilities on benchmark datasets on which the new ILUGA method has been tested. ILUGA has also shown good incremental learning ability using only a few classifiers and does not suffer from catastrophic forgetting. The results obtained for ILUGA on the Optical Character Recognition (OCR) and Wine datasets are good, with an overall accuracy of 93% and 94% respectively showing a 4% improvement over Learn++.MT for the difficult multi-class OCR dataset.

**Keywords:** Incremental Learning, Pattern Analysis and Classification, Optimization, MachineLearning, Ensemble Methods, Kernel Methods, Computational Intelligence Methods, Agroinformatics

## 1. INTRODUCTION

The general approach to making good strong classifiers is training a static classifier that cannot incorporate new data without being fully retrained. This approach of training strong static classifiers for all data is time consuming and expensive. In such situations not all the data that was previously trained is available because it has been lost or become corrupt. This makes it necessary to have a classifier that can incrementally evolve to take on novel data and classes as they become available and to not forget previously trained data. For a good incremental learning algorithm the classifier needs to be stable but with good plasticity [1]. A completely stable classifier would be able to preserve knowledge but will not be able to learn novel information, while a completely plastic classifier can learn the novel information presented to it, but cannot retain the previous knowledge. Support Vector Machines (SVMs) have shown to be stable classifiers with better pattern recognition performance than the traditional machine learning methods [2], yet stable SVM classifiers suffer from the lack of plasticity and are inclined to the catastrophic forgetting phenomenon [3, 4]. Therefore to fully benefit from the SVM classifier performance, an incremental learning method needs to be applied to the standard SVM which will retain its stability but make it plastic.

Incremental learning has been applied in different ways as shown in the literature [5]. The simplest of the incremental learning approaches is one of storing all the data which allows for retraining with all the data. At the other extreme is the training of the data, instance by instance, in an online learning fashion. Methods using the online learning approach for incremental learning have been implemented but have not considered all the issues of learning, particularly the learning of new classes [6, 7]. For a classifier to be incremental it should satisfy the following criteria:

1. *"It should be able to learn additional information from new data.*

2. *It should not require access to the original data used to train the existing classifier.*

3. *It should preserve previously acquired knowledge (that is, it should not suffer from catastrophic forgetting).*

4. *It should be able to accommodate new classes that may be introduced with new data"* [6].

Learn++ is one of the most recent incremental learning approaches that has been introduced by Polikar et al [6, 8, 9]. The approach is based on the well known AdaBoost, which uses multiple weak learning classifiers to make an incremental learning system. The approach has been modified to the Learn++.MT where it has a dynamic weight update for classes that have not been classified before [10]. Both methods have their downfalls: Learn++ performs poorly when new classes are added; Learn++.MT learns the new classes well but suffers from poor classification performance because the classes are outvoted due to the dynamic weight update, as seen with the Optical Character Recognition (OCR) dataset. Learn++, Learn++.MT, SVMLearn++ and SVMLearn++.MT will be fully described in section 4.

To solve the problems as mentioned above in Learn++ and Learn++.MT, a new incremental learning approach is developed called Incremental Learning Using Genetic Algorithm (ILUGA). The approach uses binary SVM classifiers that are trained to be strong classifiers using genetic algorithm. Once the binary classifiers are trained, genetic algorithm is used again to optimally select the weights for all the decisions of the classifiers. ILUGA will be explained in section 5.

## 2. SUPPORT VECTOR MACHINES

Support Vector Machines (SVM) were originally developed by Vapnik [11], and are based on statistical learning theory. They are margin classifiers which map the input vectors to a higher dimensional space using mapping functions $\mathbf{\Phi}(\cdot)$. From the higher dimensional space, the optimal separating hyperplane between the classes is found. SVMs try to find a classifier $f(x)$ that minimizes the misclassification rate. The classifier is implemented as $f(x) = \text{sgn}(\mathbf{w}\cdot\mathbf{x}+b)$ where the vector $\mathbf{w}$ is essentially the *kernel trick*.

### 2.1. Kernels

The kernel function is applied to maximize the margin between the hyperplanes by creating a nonlinear decision boundary. The nonlinear decision boundary set up by the kernel function allows for a further separation of the data to form a precise decision boundary. Therefore the decision of the $\mathbf{w}$ becomes an optimization problem to find the optimal separating hyperplane. The optimal kernel function is not a global best for all data; it is unique to each set of data because it finds the optimal separating plane for the input data presented [12]. Most functions can be used as kernel functions as long as they satisfy Mercer's conditions [13]. The commonly used kernel functions are linear, quadratic, radial basis function, polynomial and hyperbolic tangent.

## 3. GENETIC ALGORITHM

Genetic algorithm (GA) is a stochastic search that finds optimal solutions to problems by applying evolutionary biology such as crossover, mutation, reproduction and natural selection [14, 15]. The stochastic nature of the algorithm allows it to search in a wide range of solutions to come up with the global maximum. The GA finds the best candidates (chromosomes) by evaluating them with the fitness function which relates the optimization problem with the GA. The GA optimization consists of the following steps [14, 15]:

*1. Generate a population (pool) of possible candidates (chromosomes) for the solution.*

*2. Evaluate the fitness of each of the chromosomes, discard the chromosomes with the lowest fitness and allow only the fittest to go on to the next generation. The discarded chromosomes need to be replaced using crossover and mutation.*

*3. Points 1 and 2 are repeated until the number of generations is exceeded or a specified fitness level is obtained.*

## 4. LEARN++ METHODS

### 4.1. Learn++

Learn++ is a recent incremental learning algorithm that was introduced by Polikar et al [6, 8, 9, 10, 16]. It is based on the well known AdaBoost, which uses multiple classifiers to allow the system to learn incrementally. The algorithm works on the concept of using many classifiers that are weak learners to give a good overall classification. A weak leaner is a classifier that will classify the data with an accuracy of 50%. The weak learners are trained on a separate subset of the training data and then the classifiers are combined using a weighted majority vote. The weights for the weighted majority vote are chosen using the performance of the classifiers on the entire training dataset.

### 4.2. Learn++.MT

Learn++.MT is a modification to the Learn++ approach where the weights are dynamically updated [10]. The dynamic weight update reduces the effect of outvoting new classes as seen in Learn++ where new classes have a very low classification accuracy. It uses the technique that if an ensemble overwhelmingly chooses a class it has seen before, then the weights of the ensembles that have not seen the class, are reduced [16]. This modification shows better accuracies than those of the standard Learn++.

### 4.3. SVMLearn++ and SVMLearn++.MT
SVMLearn++ and SVMLearn++.MT are implemented in the same way as the standard Learn++ and Learn++.MT respectively except that the classifier ensemble is made up of SVMs instead of multi-layer perceptrons [16, 17].

## 5. NEW INCREMENTAL LEARNING METHOD USING GENETIC ALGORITHM
### 5.1. Overview of the algorithm
The new incremental learning approach proposed in this paper uses a different approach to the Learn++ as shown in Polikar et al [6, 8, 9, 10, 16]. Where the Learn++ approach uses weak learners to make up a large ensemble of classifiers. ILUGA uses strong SVM classifiers which are optimized. Each of the classifiers is optimized using GA to find the optimal separating hyperplane. This is done by finding the best kernel and the best soft margin. The voting weights are then generated by GA using the strong classifiers.

ILUGA applies the voting mechanism as used with many ensemble approaches whereall the classifiers vote on the class that they predict [18], except that the voting is weighted. Therefore a weighted majority voting scheme will be used. The weights that are sent to the weighted majority voting scheme are individual weights depending on the decision of the classifier as explained in section 5.3 and 5.4. ILUGA allows for as many classifiers as necessary to be trained. The training data for each new classifier that is trained is randomized so that the sections that are used for training and validating respectively will always be different, giving the classifiers new hypothesis on the data.

The algorithm is explained in four sections: the first and second section describes the training, the third section describes the voting and the fourth section describes the addition of new incremental data. The pseudo code for the algorithm is given in section 5.6.

**5.2. First stage of the training**

The first stage of training for ILUGA is where a strong classifier ensemble is built. The strong classifier ensemble is made up of binary SVM classifiers to classify the multi class dataset. The training data is randomly separated into three sections namely the training of the classifiers (**Train**), and the two validation sets (**Val1, Val2**). The binary classifiers are trained using the set **Train** which are then evolved using GA to find the optimal parameters for the SVM. The variables to be optimized are the kernel functions (quadratic, radial basis function, polynomial and hyperbolic tangent) and the soft margin. The evolution of the classifiers is done using the validation results from set **Val1** to determine the fitness of each of the chromosomes. The first stage of the training

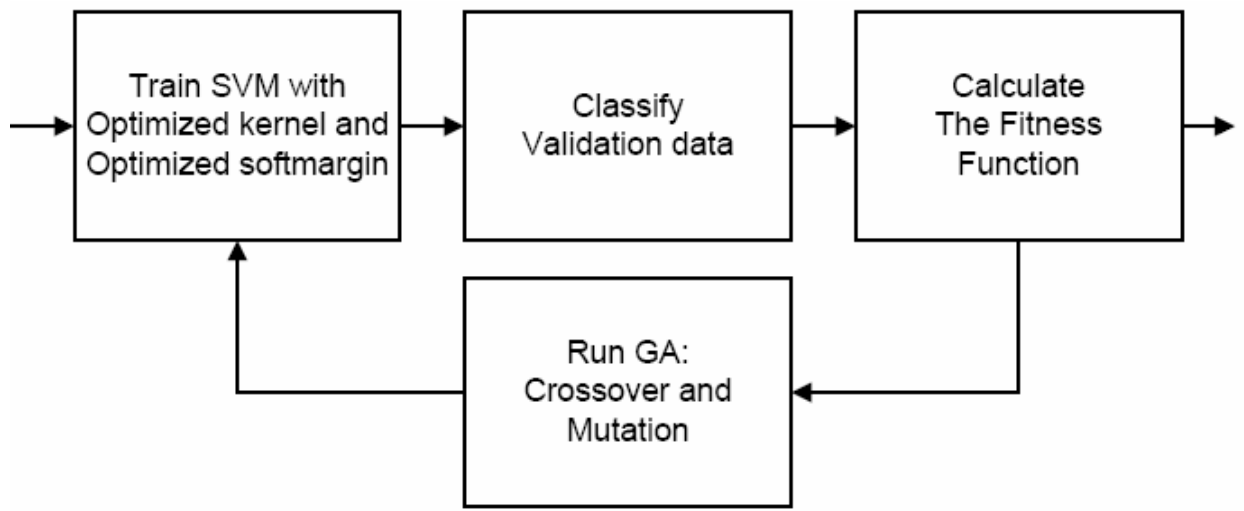

**FIGURE 1. The optimization of the SVM using genetic algorithm**

**5.3. Second stage of the training**

The second stage of training for ILUGA uses the strong binary classifier ensemble that was created in stage one. Each decision of the binary classifiers is assigned with a weight such that each binary classifier has two weights: one for each decision. The weights are then evolved using GA, which are assessed by using the fitness function. The fitness function evaluates the fitness of the chromosomes (weights) by the percentage correctly classified using weighted majority voting on set **Val2**. The optimization of individual weights makes the classification very robust thus eliminating decisions that are incorrectly classified and giving large weightings to correctly classified decisions. This also allows for new classes to be correctly classified as they are seen.

### 5.4. The voting

The voting is done to determine the class that is being classified. This is done using the ensemble of classifiers and their corresponding weights. Firstly the number of potential votes for each class is calculated, then the weights for all the classes are divided by their number of potential votes. This gives the new classes that have been introduced the same voting power as the classes that have been seen before. The weights then go into a weighted majority vote as shown in Fig. 2 where each binary classifier goes to the weighted majority vote and the highest vote is the predicted class.

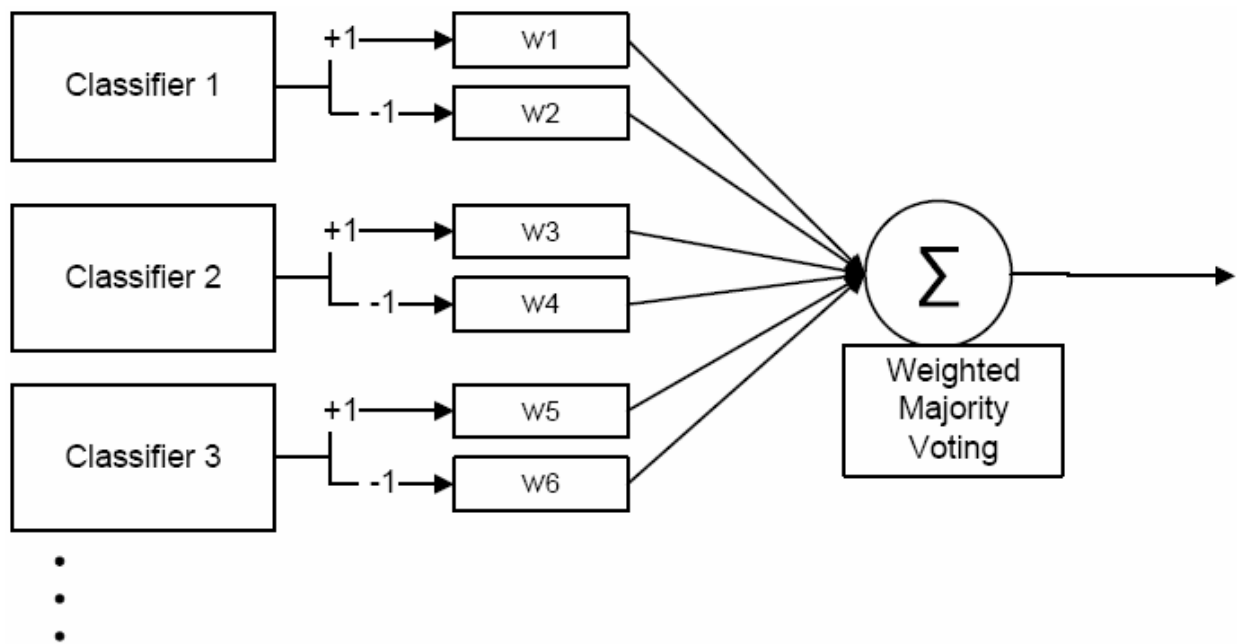

**FIGURE 2. Voting using unique weights going to the weighted majority vote**

### 5.5. Addition of new incremental data

When ILUGA is going to be incrementally trained, it needs the information of the previous classifiers, the weights and what each of the classifiers is classifying from previous training. The system then goes through stages one and two of the training, to form the new classifiers and weights which are then added to the ensemble. All the classifiers then go into the voting as shown above in section 5.4

### 5.6. Pseudo code

*Input*
- *Previous binary classifiers*
- *Previous weights*
- *Classes classified by each binary classifier*
- *Number of classifiers to be trained (m)*

*Do For* *k = 1 to m*

*1. Train strong classifier using set **Train**, then optimize the parameters using GA with set **Val1***

*2. Optimize the voting weights for each binary decision of the classifier using GA on set **Val2***

*3. Add the classifiers and weights to the ensemble*

*4. For classification, the number of potential votes for each class is calculated, then the weights for all the classes are divided by their number of potential votes. The voting is done using weighted majority voting to come up with the predicted class*

## 6. SIMULATION RESULTS

Simulation results for Learn++, Learn++.MT, SVMLearn++, SVMLearn++.MT and ILUGA were compared on the Optical Character Recognition Dataset. The top two classifiers i.e. Learn++.MT and ILUGA were then tested on the Wine Recognition Dataset. Both the benchmark datasets were obtained from the University of California, Irvine Repository of Machine Learning [19].

### 6.1. Optical character recognition dataset

The OCR Dataset has ten classes of digits 0-9 and 64 input attributes. The data was split into three sections for the training (**DS1, DS2, DS3**) and one section for the testing (**Test**). Each of the incremental learning methods were limited to the number of classifiers allowed.

SVMLearn++ and SVMLearn++.MT were allowed to train seven classifiers with the addition of each dataset. Learn++ and Learn++.MT were allowed to create five classifiers with the addition of each dataset and ILUGA was allowed to train two multi-class classifiers per dataset. The training dataset was made deliberately challenging to test the ability of the five approaches to learn multiple new classes at once and to retain the information that was previously learnt. The training dataset was set up so that each of the datasets only contain six of the ten classes that are to be trained. Classes 4 and 9 will be seen only in the final training set so that the ability of the system to incorporate new classes is fully tested. The distribution of the training and testing datasets is given in Table 1. The simulations were run many times to get a generalized average and standard deviation: Learn++, Learn++.MT and ILUGA were simulated 30 times; SVMLearn++ and SVMLearn++.MT were simulated 20 times.

Results from Learn++ and SVMlearn++ showed a generalized performance of 81% and 80% respectively, where classes 3 and 8 performed poorly after the **DS2** of the training data and classes 4 and 9 performed poorly after the **DS3** of the training data [10, 16]. The results from SVMLearn++.MT, Learn++.MT and ILUGA are shown in Table 2, 3 and 4 respectively. Each row shows the classification performance per class for the full training dataset. The last two columns show the average generalized performance (**Gen.**) and the standard deviation (**Std.**) of the generalized performance respectively.

The generalized performance was done by testing the classifier on the full dataset and not only the classes that were trained for that stage. Thus the generalized performance for sections **DS1** and **DS2** are low. The results of Learn++.MT and SVMLearn++.MT show that they learn new classes well the first time they are seen, yet using the dynamic weight update, the classes that were performing well before the new data was added are negatively effected, as though some of the classifying potential is catastrophically lost. ILUGA does not suffer from catastrophic forgetting as seen in SVMLearn++.MT and Learn++.MT and the generalized performance of ILUGA is better by over 4%. ILUGA is also using only two multi-class classifiers which makes it significantly faster than Learn++.MT and SVMLearn++.MT .

**TABLE 1. OCR data distribution**

| Class | C0 | C1 | C2 | C3 | C4 | C5 | C6 | C7 | C8 | C9 |
|---|---|---|---|---|---|---|---|---|---|---|
| DS1 | 250 | 250 | 250 | 0 | 0 | 250 | 250 | 250 | 0 | 0 |
| DS2 | 150 | 0 | 150 | 250 | 0 | 150 | 0 | 150 | 250 | 0 |
| DS3 | 0 | 150 | 0 | 150 | 400 | 0 | 150 | 0 | 150 | 400 |
| Test | 110 | 114 | 111 | 114 | 113 | 111 | 111 | 113 | 110 | 112 |

TABLE 2. SVMLearn++.MT performance results on OCR database[16]

| Class | C0 | C1 | C2 | C3 | C4 | C5 | C6 | C7 | C8 | C9 | Gen | Std |
|---|---|---|---|---|---|---|---|---|---|---|---|---|
| DS1 | 99% | 100% | 100% | - | - | 98% | 100% | 99% | - | - | 59% | 0.05% |
| DS2 | 99% | 34% | 99% | 97% | - | 93% | 20% | 99% | 60% | - | 59% | 0.43% |
| DS3 | 99% | 98% | 95% | 97% | 89% | 53% | 100% | 52% | 95% | 90% | 85% | 0.56% |

TABLE 3. Learn++.MT performance results on OCR database[10]

| Class | C0 | C1 | C2 | C3 | C4 | C5 | C6 | C7 | C8 | C9 | Gen | Std |
|---|---|---|---|---|---|---|---|---|---|---|---|---|
| DS1 | 95% | 98% | 98% | - | - | 95% | 99% | 100% | - | - | 58% | 0.8% |
| DS2 | 96% | 95% | 99% | 95% | - | 95% | 98% | 100% | 98% | - | 69% | 0.6% |
| DS3 | 67% | 95% | 92% | 98% | 83% | 63% | 98% | 100% | 95% | 96% | 89% | 0.7% |

TABLE 4. ILUGA performance results on OCR database

| Class | C0 | C1 | C2 | C3 | C4 | C5 | C6 | C7 | C8 | C9 | Gen | Std |
|---|---|---|---|---|---|---|---|---|---|---|---|---|
| DS1 | 100% | 96% | 98% | - | - | 99% | 99% | 100% | - | - | 59% | 0.2% |
| DS2 | 100% | 82% | 98% | 94% | - | 99% | 84% | 99% | 80% | - | 74% | 0.9% |
| DS3 | 92% | 97% | 96% | 92% | 95% | 90% | 99% | 98% | 90% | 83% | 93% | 1% |

6.2. Wine recognition database

The Wine Recognition dataset is then used to compare the two approaches that have performed the best on the OCR dataset namely; Learn++.MT and ILUGA. The Wine dataset has three classes with 13 input attributes. The dataset was split into two training sets (**DS1 and DS2**), a validation set (**Valid.**) and a testing set (**Test**) respectively. The distribution of the data is shown in Table 5. Both Learn++.MT and ILUGA were simulated 30 times. Using Learn++ the optimal number of classifiers were created to classify the data using the validation set with no upper limit. ILUGA used just two multi-class classifiers per section of the training data.

The results and number of classifiers trained per increment of data is shown in Table 6 and 7 for Learn++.MT and ILUGA respectively. These two approaches learn the new classes well with good generalized performance. The generalized performance for the training set **DS1** is low because it is being tested on the full testing dataset and not on just the classes that were trained. Both Learn++.MT and ILUGA show good incremental learning ability. ILUGA had only four multi-class classifiers in total, whereas the Learn++.MT approach trained eleven classifiers, making ILUGA able to classify a lot faster with better accuracy than that of Learn++.MT.

TABLE 5. Wine Recognition Dataset

| Class | C1 | C2 | C3 |
|---|---|---|---|
| DS1 | 26 | 31 | 0 |
| DS2 | 13 | 16 | 32 |
| Valid | 7 | 8 | 5 |
| Test | 13 | 16 | 11 |

TABLE 6. Learn++ MT performance on the Wine Recognition Dataset[10]

| Class | C1 | C2 | C3 | C4 | Gem | Std. |
|---|---|---|---|---|---|---|
| DS1(5 classifier | 96% | 96% | - | 70% | 70% | 6% |
| DS2(6 classifier | 99% | 87% | 90% | 92% | 92% | 5% |

## 7. DISCUSSION

On the small Wine Dataset the performance of ILUGA had only a slightly better generalized performance and better standard deviation than Learn++.MT. However with the larger dataset where there are more classes being added incrementally, ILUGA performs better with a 4% higher generalized performance than the Learn++.MT. The higher generalized performance could be explained in two ways; the first is the classifiers have optimally selected parameters, and the weights for each decision of the binary classifier use the optimization technique GA. This allows for the classifiers to classify the data with very low errors, whereas the Learn++.MT uses weak learners. The weights are also selected for a high output accuracy with unseen data, whereas Learn++.MT selects weights for an overall decision of the classifier which does not give the optimal output of the classifier.

Learn++.MT learns the new classes well but the dynamic weight update causes the classes that were classified with very high accuracies in previous training to be negatively affected and the classification of those classes are thereby decreased. This happens when a class is not trained in a certain ensemble of classifiers and the classes get outvoted because of the dynamic weight update decreasing the weights of the classifiers. ILUGA therefore benefits because there is no dynamic weight update and the weights and classifiers are chosen using an optimization technique with known data.

## 8. CONCLUSION

From the results above it can be seen that ILUGA is a robust and good incremental learning approach. ILUGA is a totally novel incremental learning approach where both the classifiers and the weights are optimized. The weights are not selected as a uniform weight for each classifier as each binary classifier is given two weights for the output decision thus helping the overall classifier to correctly predict the solution. This approach of optimizing incremental learning has been shown to perform better than Learn++.MT given the same benchmark datasets. ILUGA was also trained with only two multi-class classifiers per increment of data making it much faster with better accuracy than that of the other approaches discussed above, which used at least five new classifiers per increment of data.